\documentclass[11pt,a4paper]{article}
\usepackage{times,latexsym}
\usepackage[T1]{fontenc}
\usepackage{soul, color}
\usepackage{cjhebrew} %\ hebrew in transcript
\usepackage{float}
\usepackage{graphicx}
\usepackage{svg}
\usepackage{mathtools,amsmath,amssymb}
\usepackage{algorithmic}
\usepackage[acceptedWithA]{tacl2018v2}

\usepackage{xspace,mfirstuc,tabulary}

\newif\iftaclinstructions
\taclinstructionsfalse % AUTHORS: do NOT set this to true
\iftaclinstructions
\renewcommand{\confidential}{}
\renewcommand{\anonsubtext}{(No author info supplied here, for consistency with
TACL-submission anonymization requirements)}
\newcommand{\instr}
\fi

\iftaclpubformat % this "if" is set by the choice of options

\else

\fi
\usepackage{xspace}

\newcommand{\TOKMACRO}{{\em token-single}\xspace}
\newcommand{\MULMACRO}{{\em token-multi}\xspace}
\newcommand{\MORMACRO}{{\em morpheme}\xspace}
\newcommand{\FLIPMACRO}{{\em Hybrid}\xspace}
\newcommand{\YAPMACRO}{{\em Standard}\xspace}

\title{Neural Modeling for Named Entities and Morphology (NEMO$^2$)}

\author{
 Dan Bareket$^{1,2}$ \and Reut Tsarfaty$^1$ \\
 $^1$Bar Ilan University, Ramat-Gan, Israel\\
 $^2$Open Media and Information Lab (OMILab), The Open University of Israel, Israel\\
  {\sf dbareket@gmail.com}, {\sf reut.tsarfaty@biu.ac.il}\\
}

\date{}

\begin{document}
\maketitle
\begin{abstract}
Named Entity Recognition (NER)  is a fundamental NLP task, commonly formulated as classification over a sequence of tokens. Morphologically-Rich Languages (MRLs) pose a challenge to this basic formulation, as 
the boundaries of Named Entities do not necessarily coincide with {\em token} boundaries, rather, they respect  {\em morphological} boundaries. 
To address NER in MRLs we then need to answer two fundamental questions, namely,
what are the basic {\em units} to be labeled, and how can these units be detected and classified  in realistic settings, i.e., where no gold morphology is available.
We empirically investigate these questions on  a novel  NER benchmark, with {\em parallel} token-level and morpheme-level NER annotations, which we develop for Modern Hebrew, a morphologically rich-and-ambiguous  language. 
Our results show that 
explicitly modeling morphological boundaries  leads to improved NER performance,  and that   a novel {\em hybrid} architecture, in which NER precedes and prunes morphological decomposition, greatly outperforms the standard {\em pipeline},  where morphological decomposition  strictly precedes NER, setting a new performance bar for {\em both}  Hebrew NER and Hebrew  morphological decomposition tasks.

\end{abstract}

\section{Introduction}
\textit{Named Entity Recognition (NER)} is a fundamental task in the area of \textit{Information Extraction (IE)}, in which mentions of Named Entities (NE) 
are {\it extracted} and \textit{classified} in naturally-occurring texts.  This task is most commonly formulated as a {\em sequence labeling}   task, where  \textit{extraction}  takes the form of assigning each input token with a label that marks the {\em boundaries} of the NE (e.g., B,I,O),  and   \textit{classification} takes the form of assigning labels to  indicate   entity {\em type}   ({\sc Per, Org, Loc}, etc.). 

Despite a common initial impression from latest NER performance,  brought about by neural models on the main English NER benchmarks --- CoNLL 2003 \cite{conll2003} and OntoNotes \cite{weischedel2013ontonotes} --- the NER task in real-world settings is far from solved.
Specifically, NER performance is shown to greatly diminish when moving to other domains \cite{luan-etal-2018-multi,song2018comparison}, when addressing the long tail of rare, unseen, and new  user-generated entities \cite{derczynski-etal-2017-results}, and when handling  languages with fundamentally different structure than English.
In particular, there is no readily available and empirically verified neural modeling strategy for Neural NER in those languages with complex word-internal structure,   also known as 
{\em morphologically-rich languages}.

{\em Morphologically-rich languages} (MRL) \cite{tsarfaty-etal-2010-statistical,seddah13,tsarfaty-etal-2020-spmrl}  are languages in which substantial information concerning the arrangement of words into phrases and the relations between them  is expressed at word level, rather than in a fixed word-order or a rigid structure. The extended amount of information expressed at word-level and the  morpho-phonological processes creating these  words  result in high token-internal complexity, which poses serious challenges to the basic formulation of NER as  classification of raw, {space-delimited, tokens}.
Specifically, while NER in English is formulated as the sequence labeling of {\em space-delimited tokens}, in MRLs a single token may include multiple meaning-bearing units, henceforth {\em morphemes}, only some of which are relevant for the entity mention at hand. 

In this paper we formulate two questions concerning neural modelling strategies for NER in MRLs, namely: (i) what should be the granularity of the  units to be labeled? Space-delimited tokens or finer-grain morphological segments? and, (ii) how can we  effectively encode, and accurately detect, the morphological segments that are relevant to NER,   and specifically in {\em realistic} settings, when  gold morphological boundaries are not available?
 
To empirically investigate  these questions
we  develop  a novel {\em parallel} benchmark, containing parallel token-level and morpheme-level NER annotations for texts in Modern Hebrew --- a morphologically rich and morphologically ambiguous language,
which is known to be notoriously hard to parse \cite{more-etal-2019-joint,tsarfaty-etal-2019-whats}. 

Our results  show that morpheme-based NER is superior to token-based NER, which encourages a {\em segmentation-first} pipeline. 
At the same time, we  demonstrate that token-based NER {improves}  morphological segmentation in {\em realistic} scenarios, encouraging a {\em NER-first} pipeline. While these two findings may appear contradictory, we aim here to offer a climax;  
a {\em hybrid} architecture where the token-based NER predictions {\em precede} and  {\em prune} the space of morphological decomposition options, while the actual morpheme-based NER  takes place only {\em after} the morphological decomposition.
We empirically show that the {\em hybrid} architecture we propose outperforms all token-based and morpheme-based model variants of Hebrew NER on our benchmark, and it further outperforms all previously reported results on Hebrew NER and morphological decomposition.
Our  error analysis further demonstrates that morpheme-based models  better {\em generalize}, that is, they contribute to recognizing the long tail of  entities unseen during training ({\em out-of-vocabulary}, OOV), in particular those unseen entities that turn out to be  composed of previously {\em seen} morphemes.

The contribution of this paper is thus manifold. First, we define  key architectural questions for Neural NER modeling in MRLs and chart the space of modeling options.  Second, we deliver a new and novel parallel benchmark that allows one to {empirically} {\em compare} and {\em contrast} the morpheme vs.\ token modeling strategies. Third, we show consistent advantages for {\em morpheme-based} NER, demonstrating the importance of morphologically-aware modeling. Next we present  a novel {\em hybrid} architecture 
which demonstrates an even further improved performance on both NER and morphological decomposition tasks.
Our results for Hebrew present a new bar on these tasks, outperforming the reported state-of-the-art results on various benchmarks.\footnote{Data \& code: \scriptsize{\url{https://github.com/OnlpLab/NEMO}}}

\section{Research Questions: NER for MRLs}
\label{sec:research-questions}

In MRLs, words are internally complex, and word {\em boundaries} do not generally coincide with the boundaries of more basic  meaning-bearing units. This fact has critical ramifications for sequence labeling tasks in MRLs in general, and for NER in MRLs in particular.
Consider, for instance, the three-token Hebrew phrase in (1):\footnote{Glossing conventions are in accord with the Leipzig Glossing Rules \cite{comrie2008leipzig}.}
%\\ \scriptsize{ \url{https://www.eva.mpg.de/lingua/pdf/Glossing-Rules.pdf}, \url{http://corpafroas.huma-num.fr/fichiers/LIST_GLOSSES.pdf}}}

\begin{itemize}
    \item[(1)] \cjRL{.tsnw mt'ylnd lsyn}\\
    {\em tasnu  \hspace{0.8cm} mithailand \hspace{0.8cm} lesin}\\
    {\small flew.1PL \hspace{0.4cm} from-Thailand \hspace{0.45cm} to-China}\\
    `we flew from Thailand to China'
\end{itemize} 
It is clear that \cjRL{t'ylnd}/thailand (\emph{Thailand}) and \cjRL{syn}/sin (\emph{China}) are NEs, and in English, each NE is its own token. In the Hebrew phrase though, neither NE constitutes a single token. In either case, the NE occupies only one of two  morphemes in the token, the other being a case-assigning preposition. 
This  simple example demonstrates an extremely frequent phenomenon in MRLs such as Hebrew, Arabic or Turkish,  that the adequate boundaries for NEs do not coincide with token boundaries, and   tokens must be segmented in order to obtain accurate NE boundaries.\footnote{We use the term {\em morphological segmentation} (or {\em segmentation})  to refer to splitting raw tokens into morphological segments, each carrying a single Part-Of-Speech tag. That is, we segment away  prepositions, determiners, subordination markers and multiple kinds of pronominal clitics, that attach to their hosts via complex morpho-phonological processes.
Throughout this work, we use the terms {\em morphological segment}, {\em morpheme},  or {\em segment} interchangeably.} 

The segmentation of tokens  and the identification of adequate NE boundaries is however far from trivial, due to 
complex morpho-phonological and orthographic processes in some MRLs   \cite{vania-etal-2018-character,klein-tsarfaty-2020-getting}.
This means that the morphemes that compose  NEs are not necessarily transparent in the character sequence of the raw  tokens. Consider for example   phrase  (2):
\begin{itemize}
    \item[(2)] \cjRL{hmrw.S lbyt hlbn}
    \\ {\em hamerotz  \hspace{0.6cm} labayit  \hspace{1.2cm} halavan}
    \\ {\small the-race \hspace{0.9cm} to-house.DEF \hspace{0.3cm} the-white}
    \\ `the race to the White House'
\end{itemize} 
Here, the full form of the NE \cjRL{hbyt hlbn} / \emph{habayit halavan}  (\emph{the White House}), is not present in the utterances, only the sub-string \cjRL{byt hlbn} / \emph{bayit halavan}  (\emph{(the) White House})  
is present in (2) --- due to  phonetic and orthographic processes suppressing  the definite article \cjRL{h}/\emph{ha} in certain environments.
In this and many other cases, it is not only that NE boundaries do not coincide with token boundaries,  they do not coincide with  {\em characters} or {\em sub-strings} of the token either. This calls for accessing the more basic meaning-bearing  units of the token,   that is, to decompose the tokens into  \emph{morphemes}. 

 Unfortunately though, the morphological decomposition of surface tokens may be very  challenging due  to extreme morphological ambiguity.  
The sequence of morphemes composing a token is not always directly recoverable from its character sequence, and is not known in advance.\footnote{This ambiguity gets  magnified by the fact that Semitic languages that use abjads, like Hebrew and Arabic, lack capitalization altogether and  suppress all vowels (diacritics).} This means that for every raw space-delimited token, there are many conceivable readings which impose different segmentations, yielding different sets of potential NE boundaries.
Consider for example the token \cjRL{lbny} (\emph{lbny}) in   different contexts:

\begin{itemize}
    \item[(3)] (a) \cjRL{h/srh lbny} 
    \\ \emph{hasara \hspace{1.2cm} livni}
    \\ {\small the-minister \hspace{0.6cm} [Livni]$_{PER}$}
    \\ `Minister [Livni]$_{PER}$' 
    \\ (b) \cjRL{lbny gn.S}
    \\ {\em le-beny \hspace{1cm} gantz}
    \\ {\small for-[Benny \hspace{0.45cm} Gantz]$_{PER}$}
    \\ `for [Benny Gantz]$_{PER}$'
    \\ (c) \cjRL{lbny hyqr}
    \\ {\em li-bni \hspace{2.3cm} hayakar}
    \\ {\small for-son.POSS.1SG \hspace{0.4cm} the-dear}
    \\ `for my dear son'
    \\ (d) \cjRL{lbny xymr}
    \\ {\em livney \hspace{1.2cm} kheymar}
    \\ {\small brick.CS \hspace{0.9cm} clay}
    \\ `clay bricks'
\end{itemize} 
In (3a) the token   \cjRL{lbny} is completely consumed as a labeled NE. In (3b) \cjRL{lbny} is only partly consumed by an NE, and in (3c) and (3d) the token is entirely out of an NE context. In (3c) the token is composed of several morphemes, and in (3d) it consists of a single morpheme. These are only some of the possible decompositions of this surface token, other alternatives may still be available.
As shown by \newcite{goldberg-tsarfaty-2008-single,green-manning-2010-better,seeker-cetinoglu-2015-graph,habash-rambow-2005-arabic,more-etal-2019-joint}, and others,
the correct morphological decomposition becomes apparent only in the larger (syntactic or semantic) context. The challenge, in a nutshell, is as follows: in order to detect accurately NE boundaries, we need to segment the raw token first, however, in order to segment tokens {\em correctly}, we need to know the greater semantic content, including, e.g., the participating entities. How can we break out of this apparent loop?

Finally, MRLs are often characterized by 
an extremely sparse lexicon, consisting of a long-tail of {\em out-of-vocabulary} (OOV) entities unseen during training
\cite{czarnowska-etal-2019-dont}. Even in cases where all morphemes are present in the training data,  morphological compositions of seen morphemes 
may yield  tokens and entities which were unseen during training.  Take for example the utterance in (4), which the reader may inspect as  familiar:
\begin{itemize}
    \item[(4)] \cjRL{.tsnw msyn lt'ylnd}\\
    {\em tasnu  \hspace{0.8cm} misin \hspace{1.2cm} lethailand}\\
    {\small flew.1PL \hspace{0.4cm} from-China \hspace{0.45cm} to-Thailand}\\
    'we flew from China to Thailand'
\end{itemize} 
Example (4) is in fact example (1) with a switched flight direction. This subtle change  creates two new surface tokens \cjRL{msyn}, \cjRL{lt'ylnd} which might not have been seen during training, even if example (1) had been observed.  Morphological  compositions 
of an entity with prepositions, conjunctions, definite markers, possessive clitics and more,
cause  mentions of seen entities to have unfamiliar  surface forms, which often fail to be accurately detected and analyzed.
 
Given the aforementioned complexities, in order to solve NER for MRLs we ought to answer the following fundamental modeling questions:
  \\{\bf Q1.} {\bf  Units:}
What are the discrete units upon which we need to set NE boundaries in MRLs?
Are they  tokens?  characters? morphemes?   a representation  containing multiple levels of granularity?
  \\{\bf Q2.} {\bf Architecture:}
When employing morphemes in NER, the classical approach is  ``segmentation-first".
However, segmentation errors are detrimental and downstream NER cannot recover from them. 
How  is it best  to set up the pipeline so that segmentation and NER could interact?
  \\{\bf Q3.}  {\bf Generalization:}
How do the different modeling choices affect NER generalization in MRLs? 
How can we address the long tail of OOV NEs in MRLs? 
Which modeling strategy best handles {\em pseudo}-OOV entities that result from a {\em previously unseen}  {composition} of {\em already seen} morphemes?

\section{Formalizing NER for MRLs}
To answer the aforementioned questions, we  chart and formalize the space of  modeling options for neural NER in MRLs. 
We cast   NER   as a Sequence Labelling task and  formalize it as \(f:\mathcal{X}\rightarrow\mathcal{Y}\), where \(x\in\mathcal{X}\) is a sequence \(x_1,...,x_n\) of \(n\) discrete strings from some vocabulary \( x_i\in \Sigma\), and \(y\in \mathcal{Y}\) is a sequence \(y_1,..,y_n\) of the same length,  where \( y_i\in Labels\), and \(Labels\) is a finite set of labels composed of the BIOSE  tags (a.k.a., BIOLU as described in \citet{ratinov-roth-2009-design}).  Every non-O label is also enriched with an entity type label. Our list of types is presented in Table \ref{tab:corpus-stats}. 

\subsection{Token-Based or Morpheme-Based?}
Our first modeling question concerns the discrete units upon which to set the NE boundaries. That is, what is  the  formal definition of the input vocabulary \(\Sigma\) for the sequence labeling task?

The simplest scenario,  adopted in most NER studies,
assumes token-based input, where each token admits a single label --- hence \TOKMACRO:
\[ \textit{NER}_\textit{token-single} : \mathcal{W} \rightarrow \mathcal{L}\]
Here, \(\mathcal{W} = \{w^*| w\in \Sigma\}\) is the set of all possible token sequences in the language and  \(\mathcal{L} = \{l^*| l\in Labels\}\) is the set of all possible label sequences over the label set defined above. 
Each token gets assigned a single label, so the input and output sequences are of the same length. The drawback of this scenario is that since the input for \TOKMACRO  incorporates {\em no} morphological boundaries, the exact boundaries of the NEs remain underspecified. This case is exemplified at the top row of Table \ref{tab:model-input-output}.

There is another conceivable scenario, where the input is again the sequence of space-delimited tokens, and the output consists of complex labels (henceforth {\em multi-labels}) reflecting, for each  token, the labels of its constituent morphemes; henceforth, a \MULMACRO scenario:
\[ \textit{NER}_\textit{token-multi} : \mathcal{W} \rightarrow \mathcal{L}^*\]
Here, \(\mathcal{W} = \{w^*| w\in \Sigma\}\) is the set of sequences of  tokens as in \TOKMACRO.   
Each {\em token} is assigned a multi-label, i.e., a {sequence}  (\(l^* \in \mathcal{L}\))     which indicates the labels of the token's  morphemes in  order. The output is a sequence of such multi-labels, one multi-label per token. 
This variant incorporates  morphological information concerning the number and order of  labeled morphemes, but lacks the precise morphological boundaries. This is illustrated at the middle of Table \ref{tab:model-input-output}. A downstream application may require (possibly noisy) heuristics to determine the precise NE boundaries of each individual label in the multi-label for an input token.

Another possible scenario is a \MORMACRO-based scenario, assigning a label \(l\in L\) for each  segment:
\[ \textit{NER}_{morph} : \mathcal{M} \rightarrow \mathcal{L}\]
Here, \(\mathcal{M} = \{m^*| m\in \textit{Morphemes}\}\) is the set of  sequences of morphological segments in the language, and  \(\mathcal{L} = \{l^*| l\in \textit{Labels}\}\) is the set of label sequences as defined above. 
The upshot of this scenario is that NE boundaries are precise. An example is given in the bottom row of Table \ref{tab:model-input-output}. But, since each token may contain many meaningful morphological segments, 
the length of the  token sequence is not the same as the length of morphological segments to be labeled, and the model assumes prior morphological segmentation  --- which in realistic scenarios is not necessarily available. 

\begin{table}[t]

\scalebox{0.7}{
\begin{tabular}{|l|rl|l|}
\hline
 \textbf{Nickname} &  \textbf{Input} &  \textbf{Lit} & \textbf{Output} \\
 \hline\hline
\TOKMACRO    & \cjRL{hmrw.S} & the-race & O \\  
                  & \cjRL{lbyt}   & to-house.DEF & B\_ORG \\ 
                  & \cjRL{hlbn}   & the-white & E\_ORG  \\
                  \hline

\MULMACRO & \cjRL{hmrw.S} & the-race & O + O \\  
                  & \cjRL{lbyt}   & to-house.DEF & O + B\_ORG + I\_ORG \\ 
                  & \cjRL{hlbn}   & the-white & I\_ORG + E\_ORG \\
                  \hline

\MORMACRO    & \cjRL{h}      & the & O \\
                  & \cjRL{mrw.S}  & race & O \\  
                  & \cjRL{l}      & to & O \\  
                  & \cjRL{h}      & the & B\_ORG \\  
                  & \cjRL{byt}    & house & I\_ORG \\ 
                  & \cjRL{h}      & the & I\_ORG \\  
                  & \cjRL{lbn}    & white & E-ORG \\
\hline
\end{tabular}
}
\caption{\label{tab:model-input-output} 
Input/output for  \TOKMACRO, \MULMACRO and \MORMACRO models for example (2) in Sec.\ \ref{sec:research-questions}.
  }
\end{table}

\subsection{Realistic Morphological Decomposition} 
A major caveat with  {\em morpheme-based} modeling strategies 
is that they often assume an ideal scenario of {\em gold} morphological decomposition of the  space-delimited tokens  into morphological segments (cf.\  
\citet{nivre-etal-2007-conll, pradhan-etal-2012-conll}).
But in  reality,
 {\em gold morphological decomposition} is  {\em not} known in advance, it has to be predicted automatically, and prediction errors may  propagate to contaminate the downstream task.

Our second modeling question therefore concerns the interaction between the morphological decomposition and the NER tasks:   how would it be best to set up the pipeline so that the prediction of the two tasks can interact?

\begin{figure}[t]
\centering
  \includegraphics[width=\columnwidth]{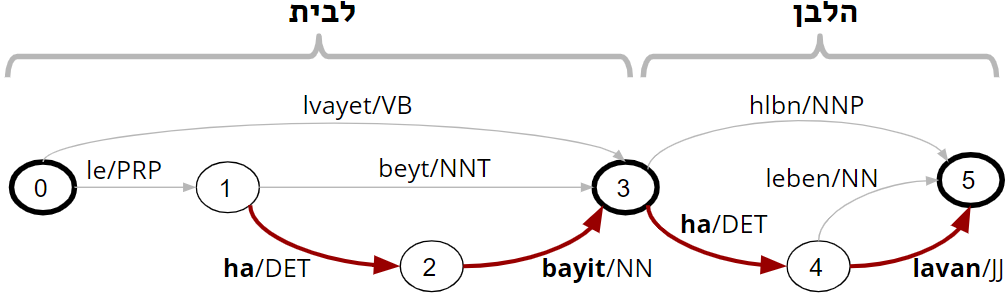}
  \caption{\label{fig:lattice} 
  Lattice for a partial list of analyses of the Hebrew tokens \cjRL{lbyt hlbn} corresponding to Table \ref{tab:model-input-output}. Bold nodes are token boundaries. Light nodes are segment boundaries. Every path through the lattice is a single morphological analysis. The bold path is a single NE.  }
\end{figure}

To answer this, we define  \emph{morphological decomposition} as consisting of two subtasks:  {\em morphological analysis} (MA) and {\em morphological disambiguation} (MD). We view sentence-based MA as:  
\[ MA : \mathcal{W} \rightarrow \mathcal{P(M)}\]
Here \(\mathcal{W} = \{w^*| w\in \Sigma\}\) is the set of  possible token sequences as before,  \(\mathcal{M} = \{m^*| m\in Morphemes\}\) is the set of possible morpheme sequences, and
\(\mathcal{P(M)}\) is the set of subsets of \(\mathcal{M}\). 
The role of \(MA\) is then to assign a token sequence  \(w\in\mathcal{W}\) with all of its {\em possible} morphological decomposition options. We represent this set of alternatives  in a dense structure  that we call a \emph{lattice} (exemplified in Figure \ref{fig:lattice}). 
MD is  the task of picking the single correct morphological path \(M\in\mathcal{M}\) through the MA lattice 
of a given sentence:
\[ MD : \mathcal{P(M)} \rightarrow \mathcal{M}\]
 
Now, assume \(x\in\mathcal{W}\) is a surface sentence in the language, with its morphological decomposition initially unknown and underspecified. 
In a  \YAPMACRO pipeline, MA strictly precedes MD: 
        \[MD_\textit{Standard}: M=MD(MA(x))\]
The main problem here is that MD errors  may propagate to contaminate the NER output.

We propose 
a novel \FLIPMACRO alternative, in which we {\em inject} a task-specific signal, in this  case NER,\footnote{We can do this for any sequence labeling task in MRLs.}  
to {\em constrain}  the search for \(M\) through the  lattice:
    \[MD_\textit{Hybrid}: M=MD(MA(x)\restriction
     \textit{NER}_\textit{token}(x))\]
Here, the restriction \(MA(x)\restriction NER(x)\) indicates {\em pruning} the  lattice structure \(MA(x)\) to  contain {\em only}  MD options that are  {\em compatible} with the token-based NER  predictions, and only then apply \(MD\) to the pruned lattice.

Both \(MD_\textit{Standard}\) and \(MD_\textit{Hybrid}\) are  disambiguation architectures that result in a  morpheme sequence \(M\in\mathcal{M}\). The latter benefits from the NER signal, while the former doesn't.
The sequence \(M\in\mathcal{M}\)  can  be used in one of two ways. 
 We can use  \(M\) as \emph{input} to a \MORMACRO model to output morpheme labels.
Or, we can {rely on} the {\em output} of the  \MULMACRO model and align the token's multi-label with the  segments  in \(M\).

In what follows, we want to empirically assess the effect of different modeling choices (\TOKMACRO, \MULMACRO, \MORMACRO) and disambiguation architectures ({\em Standard, Hybrid}) on the performance of NER in MRLs. To this end, we need a corpus that allows  training and evaluating NER  at both  token and morpheme-level granularity.

\section{The Data: A Novel NER Corpus} 
\label{sec:the-data}

This work empirically investigates NER modeling strategies in  Hebrew, a Semitic language known for its complex and highly ambiguous morphology.
 \citet{naama}, the only previous work on Hebrew NER to date, annotated  space-delimited tokens, basing their guidelines on the CoNLL 2003 shared task \cite{Chinchor99}.  
 
 Popular Arabic NER corpora also label space-delimited tokens (ANERcorp \cite{benajiba-2007-anersys}, AQMAR \cite{mohit-etal-2012-recall}, TWEETS \cite{darwish-2013-named}), with the exception of the Arabic portion of OntoNotes \cite{weischedel2013ontonotes} and ACE \cite{linguistic2008ace} 
which annotate NER labels on gold morphologically pre-segmented texts. However, these works do not provide  a  comprehensive analysis  on the performance gaps between \MORMACRO-based and {\em token}-based  scenarios. 

In  agglutinative languages  as Turkish, token segmentation is {\em always} performed before NER (\citet{tur-data-2003,turk-tweet-seker}, re-enforcing the need to contrast the  {\em token}-based scenario, widely adopted for Semitic languages, with the \MORMACRO-based scenarios in other MRLs. 

Our  first contribution is thus a   {\em parallel} corpus for Hebrew NER, one version consists of gold-labeled tokens and the other consists of gold-labeled morphemes, for the same text. 
For this, we performed gold NE annotation of the Hebrew Treebank \cite{simaan2001}, based on the 6,143  morpho-syntactically analyzed sentences of the HAARETZ corpus, to create both token-level and morpheme-level variants, as illustrated   at the topmost and lowest rows of Table \ref{tab:model-input-output}, respectively.

\paragraph{Annotation Scheme} We  started off with the guidelines of \citet{naama}, from which we deviate in three main ways. First, 
we label NE boundaries and their types on sequences of {\em morphemes}, 
in addition to the space-delimited token annotations.\footnote{A single NE is always continuous. Token-morpheme discrepancies do not lead to discontinuous NEs.}
Secondly, we use the  finer-grained entity categories list of ACE \cite{linguistic2008ace}.\footnote{Entity categories are listed in Table \ref{tab:corpus-stats}.
We dropped the \textsc{NORP} category, since it introduced complexity concerning the distinction between adjectives and group names. \textsc{Law} did not appear in our corpus.}
Finally, we allow {\em nested} entity mentions, as in  \citet{Finkel:2009:NNE:1699510.1699529,benikova-etal-2014-nosta}.\footnote{Nested labels are  are not modeled in this paper, but they are published with the corpus, to allow for further research.}

\paragraph{Annotation Cycle}
As  \citet{fort-etal-2009-towards} put it,
examples and rules would never cover all possible cases because of the specificity of natural language and the ambiguity of formulation.
To address this we employed the cyclic approach of 
{\em agile}  annotation as offered by \citet{alex-etal-2010-agile}. Every cycle consisted of: {\em annotation}, {\em evaluation and  curation}, {\em clarification and  refinements}.  We used WebAnno \cite{yimam-etal-2013-webanno} as our annotation interface. 

{\em The Initial Annotation Cycle} was  a two-stage pilot with 12 participants, divided into 2 teams of 6. 
The teams received the same guidelines, with the exception of the specifications of entity boundaries. One team was guided to annotate the minimal string that designates the entity. The other was guided to tag the maximal string which can still be considered as the entity. Our agreement analysis showed that the minimal guideline generally led to more consistent annotations.
Based on this result (as well as low-level refinements) from the pilot,  we devised the  full version of the guidelines.\footnote{The complete annotation guide is publicly available at \scriptsize{\url{https://github.com/OnlpLab/NEMO-Corpus}}.} 

{\em Annotation,} {\em Evaluation and Curation:} 
Every annotation cycle was performed by two annotators (\emph{A}, \emph{B}) and an annotation manager/curator (\emph{C}). We annotated the full corpus in 7 cycles.
 We evaluated the annotation  in two ways, manual curation and automatic evaluation.
After each annotation step, the curator manually reviewed every sentence in which disagreements arose, as well as specific points of difficulty pointed out by the annotators. The inter-annotator agreement metric described  below was also used to quantitatively gauge the progress and quality of the annotation.

{\em Clarifications and Refinements:} In the end of each cycle we held a clarification talk between A, B and C, 
in which issues that came up during the cycle were discussed.
Following that talk we refined the guidelines and updated the annotators, which went on to the next cycle. 
In the end we performed a final curation run to make sentences from earlier cycles comply with later refinements.\footnote{A, B and C annotations are published to enable research on learning with disagreements \cite{plank-etal-2014-learning}.}

{\bf Inter-Annotator Agreement (IAA)}
\label{subsec:iaa}
IAA is commonly measured using  $\kappa$-statistic. However, \citet{pyysalo2007bioinfer} show that it is not suitable for evaluating inter-annotator agreement in NER. Instead, an $F_1$ metric on entity mentions has  in recent years been adopted for this purpose \cite{zhang2013named}. 
This metric allows for computing pair-wise IAA using standard $F_1$ score
by treating one annotator as gold and the other as the prediction.

Our full corpus pair-wise \(F_1\) scores are: IAA(\emph{A},\emph{B})=89, IAA(\emph{B},\emph{C})=92, IAA(\emph{A},\emph{C})=96. Table \ref{tab:corpus-stats} presents final corpus statistics.

{\bf Annotation Costs} The annotation took on average about 35 seconds per sentence, and thus a total of 60 hours for all sentences in the corpus for each annotator. Six clarification talks were held between the cycles, which lasted from thirty minutes to an hour. Giving a total of about 130 work hours of expert annotators.\footnote{The corpus is available at \scriptsize{\url{https://github.com/OnlpLab/NEMO-Corpus}}.}  
\label{subsec:costs}

\begin{table}[t]
\centering
\scalebox{0.7}{
\begin{tabular}{|l|lll|}
\hline
 &  \textbf{train} &   \textbf{dev} &  \textbf{test} \\\hline
 \hline
\textbf{Sentences} & $4,937$ & $500$ & $706$ \\
\textbf{Tokens} & $93,504$ & $8,531$ & $12,619$ \\
\textbf{Morphemes} & $127,031$ & $11,301$ & $ 16,828$ \\
\hline
\hline
\textbf{All mentions} & $6,282$ & $499$ & $932$ \\
\hline\hline
{\bf Type:} \text{Person \sc{(Per)}} & $2,128$ & $193$ & $267$ \\
{\bf Type:} \text{Organization} \sc{(Org)} & $2,043$ & $119$ & $408$ \\
{\bf Type:} \text{Geo-Political} \sc{(Gpe)} & $1,377$ & $121$ & $195$ \\
{\bf Type:} \text{Location} \sc{(Loc)} &  $331$ &  $28$ &  $41$ \\
{\bf Type:} \text{Facility} \sc{(Fac)} &  $163$ &  $12$ &  $11$ \\
{\bf Type:} \text{Work-of-Art} \sc{(Woa)} &  $114$ &   $9$ &   $6$ \\
{\bf Type:} \text{Event} \sc{(Eve)} &   $57$ &  $12$ &   $0$ \\
{\bf Type:} \text{Product} \sc{(Duc)} &   $36$ &   $2$ &   $3$ \\
{\bf Type:} \text{Language} \sc{(Ang)} &   $33$ &   $3$ &   $1$ \\
\hline
\end{tabular}
}
\caption{\label{tab:corpus-stats} 
Basic Corpus Statistics. Standard HTB Splits.}
\end{table}

\section{Experimental Settings}
\paragraph{Goal}
We set out to empirically evaluate the representation alternatives for the input/output sequences (\TOKMACRO, \MULMACRO, \MORMACRO)  and the effect of  different    architectures (\YAPMACRO,  \FLIPMACRO) on  the  performance  of  NER  for  Hebrew.

\paragraph{Modeling Variants}
All experiments  use the corpus we just described  and employ a standard Bi-LSTM-CRF  architecture  for implementing the  neural sequence labeling task \cite{DBLP:journals/corr/HuangXY15}.
Our basic architecture\footnote{Using the NCRF++ suite of    \citet{yang2018ncrf}.} is composed of an embedding layer for the input 
and a 2-layer Bi-LSTM followed by a CRF inference layer --- for which we test three modeling variants.

\begin{figure}[t]
\centering
  \includegraphics[width=\columnwidth]{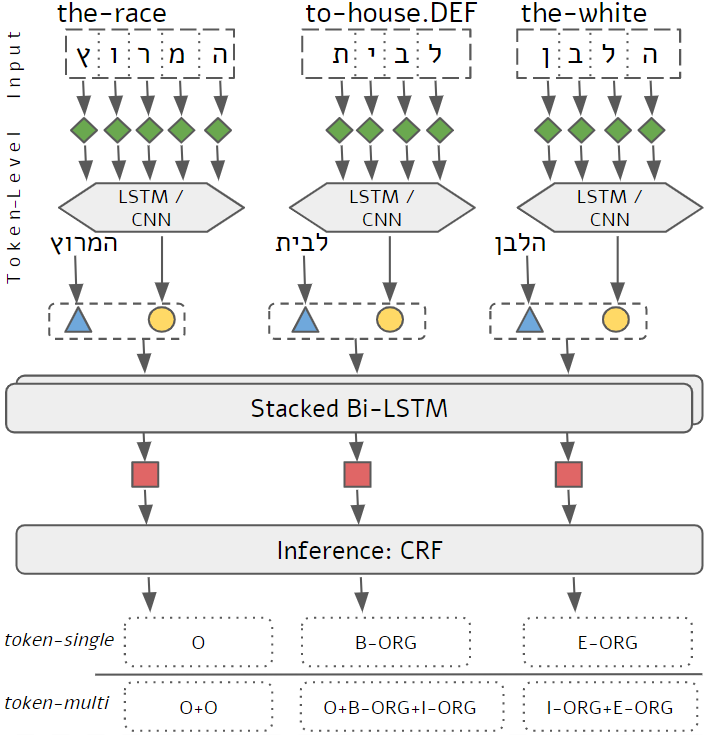}
  %\def\svgwidth{\columnwidth}
  %\includesvg{images/37_LTR_TOK}
  \caption{\label{fig:token-model} 
  The \TOKMACRO and \MULMACRO Models. The input and output correspond to rows 1,2 in Tab.\ \ref{tab:model-input-output}. Triangles indicate   {\em string} embeddings. Circles indicate   {\em char-based} encoding.}
\end{figure}

\begin{figure}[t]
\centering
  \includegraphics[width=\columnwidth]{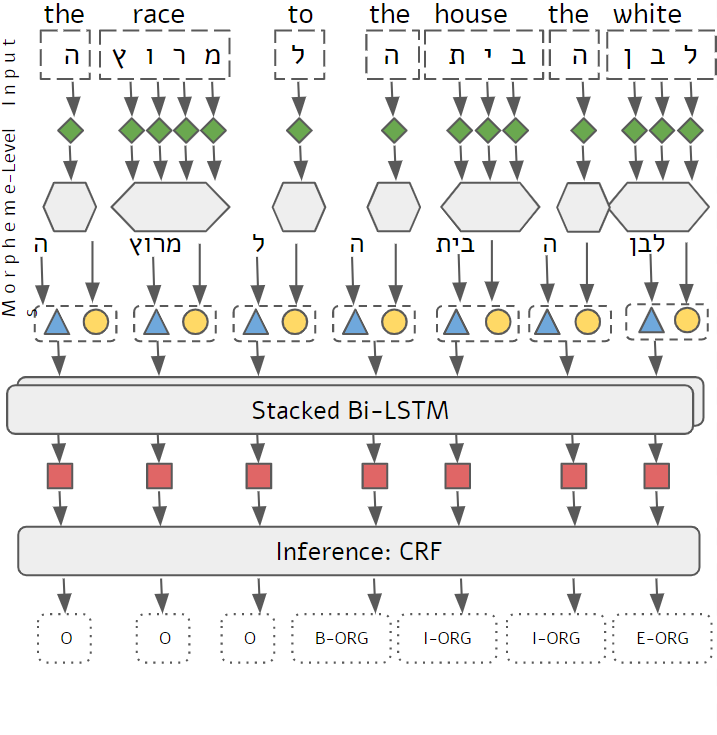}
  %\def\svgwidth{\columnwidth}
  %\includesvg{images/37_LTR_MORPH}
  \caption{\label{fig:morph-model} 
  The \MORMACRO Model. The input and output correspond to row 3 in Tab.\ \ref{tab:model-input-output}. Triangles indicate  {\em string} embeddings. Circles indicate   {\em char-based} encoding.}
\end{figure}

Figures \ref{fig:token-model}--\ref{fig:morph-model} present  the   variants we employ. 
Figure \ref{fig:token-model} shows   the token-based variants, \TOKMACRO and \MULMACRO.
The former outputs a single BIOSE  label per  token, and the latter 
 outputs a multi-label per token --- a concatenation of 
 BIOSE labels of the morphemes composing the  token. Figure \ref{fig:morph-model} shows the \MORMACRO-based variant for the same input phrase. It has the same basic architecture, but now the input consists of  morphological segments instead of tokens.  The model outputs a single BIOSE label for each morphological segment in the input.

In all modeling variants, the input may be encoded in two  ways: 
(a) String-level embeddings (token-based or morpheme-based) optionally initialized with pre-trained embeddings.
(b) Char-level embeddings, trained simultaneously with the main task 
(cf.\ \citet{DBLP:journals/corr/MaH16,DBLP:journals/corr/ChiuN15,DBLP:journals/corr/LampleBSKD16}). 
For char-based encoding (of either tokens or morphemes) we experiment with CharLSTM,  CharCNN or NoChar, that is, no character embedding at all.

We pre-trained all  token-based or morpheme-based embeddings on the  Hebrew Wikipedia dump of \newcite{wikipedia-yoav-2014}. For morpheme-based embeddings, 
we  decompose the input using  \newcite{more-etal-2019-joint}, and use the morphological segments as the embedding units.\footnote{Embeddings and Wikipedia corpus also available in: \scriptsize{\url{https://github.com/OnlpLab/NEMO}}} 
We compare GloVe \cite{pennington-etal-2014-glove} and fastText \cite{bojanowski2017enriching}. 
We hypothesize  that since FastText uses sub-string information, it will be more useful for analyzing OOVs.

 \paragraph{Hyper parameters}
Following \newcite{DBLP:journals/corr/ReimersG17,yang2018design}, we performed hyper-parameter tuning for each of our model variants. We performed hyper-parameter tuning on the dev set in a number of rounds of random search, independently on every input/output and char-embedding architecture. 
Table \ref{tab:hyperparams} shows our selected hyper-parameters.\footnote{A few interesting
empirical observations diverging from those of \citet{DBLP:journals/corr/ReimersG17,yang2018design} are worth mentioning. 
We found that a lower {\em Learning Rate} than the one  recommended by \citet{yang2018design} (0.015), led to better results and less occurrences of divergence.
We further found that raising the number of {\em Epochs} from 100 to 200 did not result in over-fitting, and significantly improved NER results. We used for evaluation the weights from the best epoch.}
The {\em Char CNN window size} is particularly interesting as it was not treated as a hyper-parameter in  \citet{DBLP:journals/corr/ReimersG17}, \citet{yang2018design}. 
However, given the token-internal complexity in MRLs we conjecture that the window size over characters might make a crucial effect. In our experiments we found that  a larger window (7)  increased the performance. For MRLs,  further research into this  hyper-parameter  might be of interest.

\begin{table}[t]
\scalebox{0.8}{

\begin{tabular}{|l|l||l|l|}

\hline
Parameter & Value & Parameter & Value \\
\hline
\hline
 Optimizer & SGD & *LR (\TOKMACRO) & 0.01 \\ 
*Batch Size & 8  & *LR (\MULMACRO) & 0.005 \\ 
LR decay & 0.05 & *LR ({\MORMACRO}) & 0.01\\
Epochs & 200 & Dropout & 0.5 \\
Bi-LSTM layers & 2 & *CharCNN window & 7 \\
*Word Emb Dim & 300 &  Char Emb dim & 30 \\
Word Hidden Dim & 200 & *Char Hidden Dim & 70  \\

\hline
\end{tabular}
}
\caption{\label{tab:hyperparams} 
Summary of Hyper-Parameter Tuning. The * indicates divergence from the NCRF++  proposed setup and empirical findings \cite{yang2018ncrf}.
  }
\end{table}

\paragraph{Evaluation} 

Standard NER studies typically invoke the CoNLL evaluation script that anchors NEs in token {\em positions}  \cite{conll2003}. However, it is inadequate for our purposes because we want to compare entities across token-based vs.\ morpheme-based settings. To this end, we use a  revised evaluation procedure, which anchors the entity in its {\em form} rather than its {\em index}. 
Specifically, we report \(F_1\) scores on strict, exact-match of the surface forms of the entity mentions. I.e.,  the gold and predicted NE spans must exactly match in their form, boundaries, and  entity type. 
In all experiments, we report   both token-level F-scores and morpheme-level F-scores, for all models. 
\begin{itemize}
\item  {\bf Token-Level evaluation.} 
For the sake of backwards compatibility with  previous work on Hebrew NER, we first define {\em token-level}  evaluation. 
    For \TOKMACRO this is a straightforward calculation of \(F_1\)  against gold spans. 
    For \MULMACRO and \MORMACRO, we need to map the predicted label sequence of that token to a single label, and we do so using linguistically-informed rules we devise
    (as elaborated in Appendix \ref{sec:appendix-extending-morph}).\footnote{In the \MORMACRO case we might encounter ``illegal'' label sequences in case of a prediction error.  We employ similar linguistically-informed heuristics to recover from that (See Appendix \ref{sec:appendix-extending-morph}).}

\item {\bf Morpheme-Level evaluation.}
 Our ultimate goal is to obtain {\em precise}   boundaries of the NEs.
 Thus, our main metric evaluates  NEs against the {gold} {\em morphological}  boundaries.  
  For \MORMACRO and \TOKMACRO models, this is a straightforward \(F_1\) calculation against gold spans. Note for  \TOKMACRO we are expected to pay a price for boundary mismatch.
  For \MULMACRO, we  know the number and order of labels, so we align the labels  in the  multi-label of the token with the morphemes in its morphological decomposition.\footnote{In  case of a misalignment  (in the number of morphemes and  labels) we match the label-morpheme pairs from the final one  backwards, and pad unpaired morphemes with O labels.}

\end{itemize}
For all experiments and metrics, we report mean and confidence interval (0.95) over ten runs.

\paragraph{Input-Output Scenarios}
We experiment with two kinds of input settings:
{\em token}-based, where the input consists of the sequence of space-delimited tokens, and \MORMACRO-based, where the input consists of  morphological segments. For the \MORMACRO input, there are three input variants:
    \begin{itemize}
    \item[] 
     {\em (i) Morph-gold}:  where the  morphological sequence is produced by an expert (idealistic). 
     \\ {\em (ii) Morph-standard}:  where the  morphological sequence is produced by a standard segmentation-first pipeline (realistic).
    \\ {\em (iii) Morph-hybrid}:  where the  morphological sequence is produced by the hybrid architecture we propose (realistic).
   \end{itemize}
     
In the \MULMACRO case we can perform {\em morpheme-based evaluation} by aligning individual labels in the multi-label with the morpheme sequence of the respective token.  Again we have three options as to which morphemes to use:
 
     \begin{itemize}
      \item[] {\em (i) Tok-multi-gold}:   The multi-label is aligned with morphemes produced by an expert (idealistic). 
    \\ {\em (ii) Tok-multi-standard}:   The  multi-label is aligned with morphemes produced by a standard pipeline (realistic).
  \\ {\em (iii) Tok-multi-hybrid}:   The multi-label  is aligned with morphemes produced by the hybrid architecture we propose (realistic).
\end{itemize} 

\paragraph{Pipeline Scenarios}
Assume an input sentence \(x\). In the \YAPMACRO
pipeline we use YAP,\footnote{For other languages this may be done using  models for canonical segmentation 
as in \cite{kann-etal-2016-neural}.} the current state-of-the-art morpho-syntactic  parser for  Hebrew \cite{more-etal-2019-joint}, for the predicted segmentation  \(M=MD(MA(x))\). 
In the \FLIPMACRO pipeline, we use YAP to first generate complete morphological lattices  
\(MA(x)\). Then, to obtain \(MA(x)\restriction NER(x)\) we omit lattice paths where the number of morphemes in the token decomposition  does not conform with the number of labels in the multi-label  of \(\textit{NER}_\textit{\MULMACRO}(x)\).
Then, we apply YAP to obtain \(MD(MA(x)\restriction NER(x))\) on the constrained lattice.
In predicted morphology scenarios (either \YAPMACRO or \FLIPMACRO), we use the same model weights  as  trained on the gold segments, but feed predicted morphemes as input.\footnote{We do not re-train the \MORMACRO models with predicted segmentation, which might achieve better performance (e.g.. jackknifing). We leave this for future work.}

\section{Results}

\subsection{The Units: Tokens vs.\ Morphemes}
\label{sec:units}

\begin{figure}[t]\scalebox{0.9}{
  \includegraphics[width=\columnwidth]{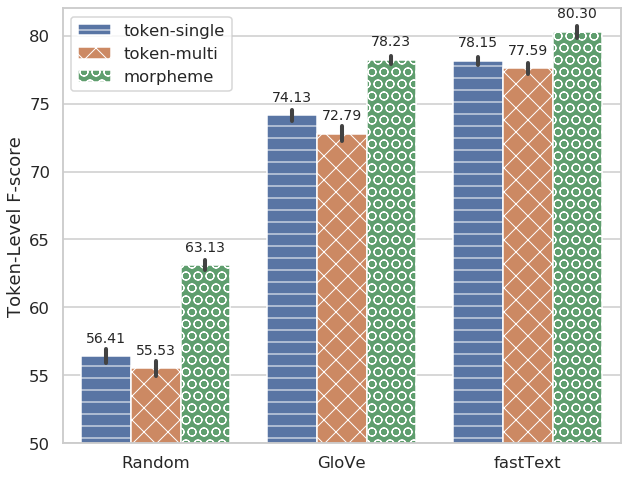}}
  \caption{\label{fig:tok-morph-tokeval} 
  Token-level Eval.\ on Dev w/ Gold Segmentation. CharCNN for morph, CharLSTM for tok.}
\end{figure}
\begin{figure}[t]\scalebox{0.9}{
  \includegraphics[width=\columnwidth]{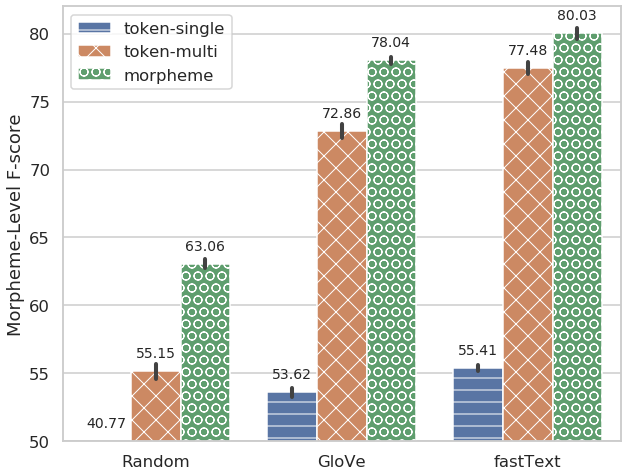}}
  \caption{\label{fig:tok-morph-morpheval}
  Morph-Level Eval.\ on Dev w/ Gold Segmentation. CharCNN for morph, CharLSTM for tok.}
\end{figure}

\begin{figure}[t]\scalebox{0.9}{
  \includegraphics[width=\columnwidth]{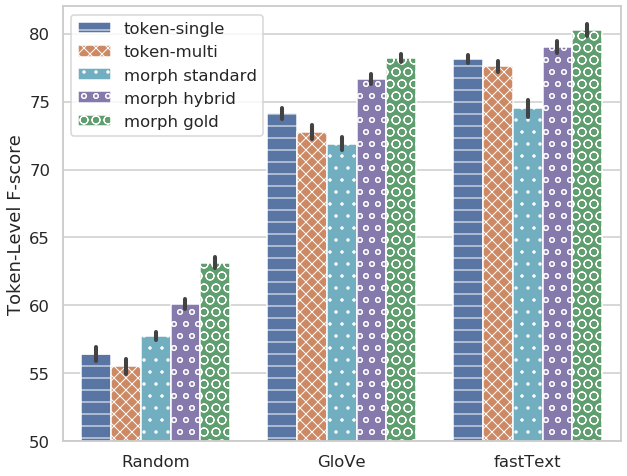}}
  \caption{\label{fig:realistic-tokeval} 
  Token-Level Evaluation in  Realistic Scenarios on Dev, comparing \emph{Gold}, \YAPMACRO and \FLIPMACRO   Morphology. CharCNN for morph, CharLSTM for tok. Results for {\em Gold}, \TOKMACRO and \MULMACRO are taken from Fig~\ref{fig:tok-morph-tokeval}. }
\end{figure}

\begin{figure}[t]\scalebox{0.9}{
  \includegraphics[width=\columnwidth]{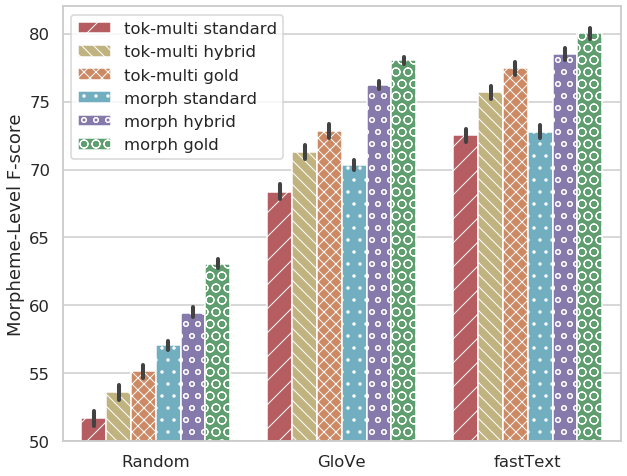}}
  \caption{\label{fig:realistic-morpheval} 
 Morph-Level Evaluation in  Realistic Scenarios
 on Dev, comparing \emph{Gold}, \YAPMACRO and \FLIPMACRO   Morphology. CharCNN for morph, CharLSTM for tok. Results for {\em Gold}, \TOKMACRO and \MULMACRO are taken from Fig~\ref{fig:tok-morph-morpheval}.}
\end{figure}

Figure \ref{fig:tok-morph-tokeval} shows the token-level evaluation for the different model variants we defined. We see that \MORMACRO models perform significantly better than the \TOKMACRO and \MULMACRO variants. Interestingly,  explicit modeling of morphemes leads to better NER performance even when evaluated against token-level boundaries. As expected, the performance gaps between variants are smaller with fastText  than they are with  embeddings that are unaware of characters (GloVe) or with no pre-training at all. We further pursue 
this in Sec.~\ref{subsec:oov}.

Figure \ref{fig:tok-morph-morpheval} shows the morpheme-level evaluation for the same model variants as in Table \ref{fig:tok-morph-tokeval}. The most obvious trend here is the drop in the performance of the \TOKMACRO model. This is expected, reflecting the inadequacy of token boundaries for  identifying accurate boundaries for NER. Interestingly, \MORMACRO and \MULMACRO models keep a  similar level of performance as in token-level evaluation, only  slightly lower. Their performance gap is  also maintained,  with \MORMACRO performing better than \MULMACRO. An obvious caveat is that these results are obtained with {\em gold} morphology. What happens in realistic scenarios?

\subsection{The Architecture: Pipeline vs.\ Hybrid}
\label{sec:architectures-pipeline-hybrid}

\begin{figure}[t]\scalebox{0.9}{
  \includegraphics[width=\columnwidth]{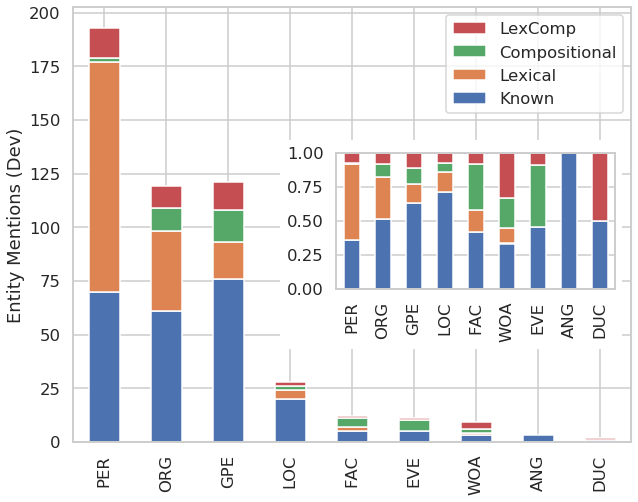}}
  \caption{\label{fig:analysis_ment_cat_ootv} 
  Entity Mention Counts and Ratio by Category and OOTV Category, for Dev Set.}
\end{figure}

Figure \ref{fig:realistic-tokeval} shows  the token-level evaluation results in realistic  scenarios. We first observe a significant drop for \MORMACRO models when \YAPMACRO predicted  segmentation is introduced instead of {\em gold}.   This means that MD errors are indeed detrimental for the downstream task, in a non-negligible rate. Second, we observe that much of this performance gap  is recovered with the \FLIPMACRO pipeline. 
It is noteworthy that while {\em morph hybrid} lags behind {\em morph gold}, it is still consistently better  than token-based models, \TOKMACRO and \MULMACRO.

Figure \ref{fig:realistic-morpheval} shows morpheme-level evaluation results for the same scenarios as in Table~\ref{fig:realistic-tokeval}. 
All trends from the token-level evaluation persist, including  a drop for all models with predicted segmentation relative to {\em gold}, with the {\em hybrid} variant recovering much of the gap.
Again {\em morph gold}  outperforms \MULMACRO, but {\em morph hybrid} shows great advantages over {\em all} {\em tok-multi} variants. This  performance gap  between {\em morph} ({\em gold} or {\em hybrid}) and {\em tok-multi} indicates that {\em explicit} morphological modeling is indeed crucial for  accurate NER.

\begin{figure}[t]\scalebox{0.9}{
  \includegraphics[width=\columnwidth]{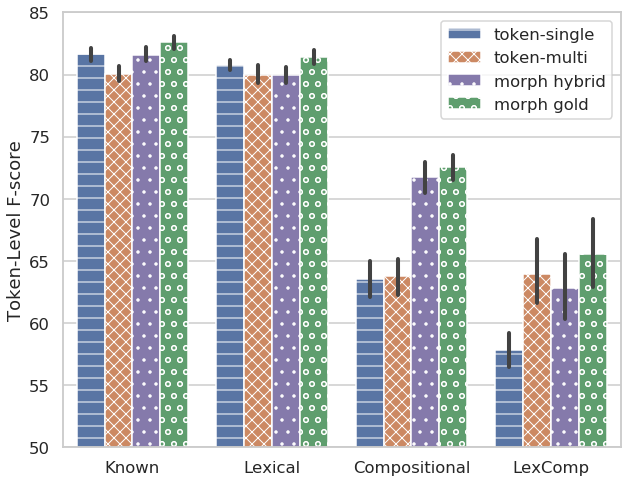}}
  \caption{\label{fig:analysis-ootv-fasttext} 
  Token-Level Eval on Dev by OOTV Category. Using fastText and CharLSTM.}
\end{figure}

\subsection{Morphologically-Aware OOV Evaluation}
\label{subsec:oov}

As discussed in Section \ref{sec:research-questions}, morphological composition introduces an extremely sparse word-level ``long-tail'' in MRLs. In order to gauge this phenomenon and its effects on NER performance, we categorize unseen, out-of-training-vocabulary (OOTV) mentions into 3 categories:
\begin{itemize}
    \item \emph{Lexical}: Unknown mentions caused by an unknown token which consists of a single morpheme. This is a strictly lexical unknown with no morphological composition  (most English   unknowns are in this category).
    \item \emph{Compositional}: Unknown mentions caused by an unknown token which consists of multiple \emph{known} morphemes. These are unknowns introduced strictly by morphological composition, with no lexical unknowns.
    \item \emph{LexComp}: Unknown mentions caused by an unknown token consisting of multiple morphemes, of which (at least) one morpheme was not seen during training. In such cases,  both unknown morphological composition \emph{and} lexical unknowns are involved. 
\end{itemize}

We group NEs based on these categories, and evaluate each group separately. We consider mentions that do not fall into any category as \emph{Known}.

Figure \ref{fig:analysis_ment_cat_ootv} shows the distributions of entity mentions in the dev set by entity type and OOTV category.
OOTV categories that involve composition ({\em Comp} and {\em LexComp}) are spread across all  categories but one, and in some they even make up more than half of all mentions.

Figure \ref{fig:analysis-ootv-fasttext} shows token-level evaluation\footnote{This section focuses on token-level  evaluation, which is a  permissive evaluation metric,
allowing us to compare the models on a more level playing field, where all models (including \TOKMACRO) have an equal opportunity to perform.}  with fastText embeddings, grouped by OOTV type. We first observe that indeed unknown NEs that are due to  morphological composition (\emph{Comp} and \emph{LexComp}) proved the most challenging for {all} models. We also find that in strictly \emph{Compositional} OOTV mentions, \MORMACRO-based models exhibit their most significant performance advantage, supporting the hypothesis that explicit {morphology} helps to {\em generalize}.  We finally observe that \MULMACRO models perform better than \TOKMACRO models for these NEs (in contrast with the trend for non-compositional  NEs). This  corroborates the hypothesis that even partial modeling of morphology (as in  \MULMACRO compared to \TOKMACRO) is better than none, leading to better generalization.

\begin{figure*}[t]
  \includegraphics[width=1.0\textwidth]{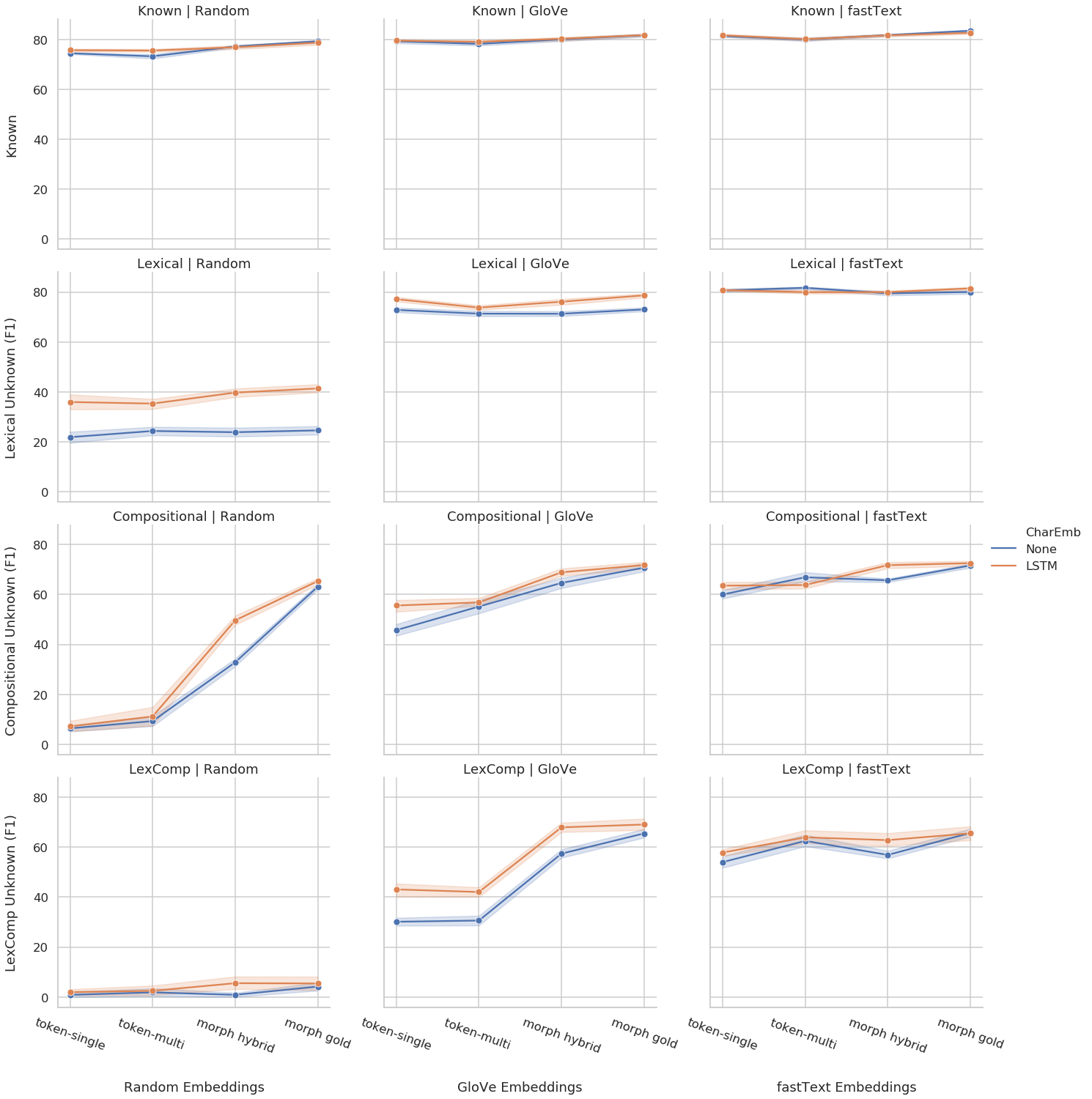}
  \caption{\label{fig:analysis_char} 
  Token-Level Eval.\ on Dev for Different OOTV Types, Char- and Word-Embeddings.}
  %Area around line signifies confidence interval 0.95 
\end{figure*}
\paragraph{String-level vs.\ Character-level Embeddings} 

To further understand the generalization capacity of different modeling alternatives in MRLs, we probe into the interplay of string-based and char-based embeddings in treating OOTV NEs. 

Figure \ref{fig:analysis_char} presents 12 plots, each of which  presents  the level of performance (y-axes)  for all models (x-axes). Token-based models are on the left of each x-axes, morpheme-based are on the right.  We plot results with and without character embeddings,\footnote{For brevity we only show char LSTM (vs.\ no char representation), there was no significant difference with CNN.} in orange and blue respectively. The plots are organized in a large grid, with the type of  NE  on the y-axes ({\em Known}, {\em Lex}, {\em Comp}, {\em LexComp}), and the type of  pre-training on the x-axes (No pre-training, GloVe, fastText) .  

At the top-most row, plotting the accuracy for {\em Known} NEs, we see a high level of performance for all pre-training methods, with not much differences between the type of  pre-training, with or without the  character embeddings. Moving further down to the row of  {\em Lexical} unseen NEs, char-based representations lead to significant advantages when we assume no pre-training, but with GloVe pre-training the performance substantially increases, and with fastText the differences in performance with/without char-embeddings almost entirely diminish, indicating the char-based embeddings are somewhat redundant in this case.

The two lower rows in the large  grid show the performance for {\em Comp} and {\em LexComp} unseen NEs,   which are ubiquitous in MRLs. For {\em Compositional} NEs,  pre-training closes only part of the gap between token-based and morpheme-based models. Adding char-based representations indeed {helps} the token-based models, but crucially does {\em not} close the gap with the morpheme-based variants.   

Finally, for {\em LexComp} NEs at the lowest row, we again see that adding GloVe pre-training and char-based embeddings does not close the gap with morpheme-based models, indicating that not all morphological information is  captured by these vectors. For fastText with char-based embeddings the gap between \MULMACRO  and \MORMACRO greatly diminishes, but is still well above \TOKMACRO. This suggests 
 biasing the model to learn about morphology (either via multi-labels or by incorporating morphological boundaries) has advantages for analysing OOTV entities, beyond the contribution of char-based embeddings alone.

All in all, the biggest advantage of
morpheme-based models over token-based models is their ability to generalize from observed tokens to composition-related OOTV ({\em Comp/LexComp}).
While character-based embeddings do help token-based models  generalize, the contribution of modeling morphology is indispensable, above and beyond the  contribution of char-based embeddings.

\begin{table}[t]
\centering
\scalebox{0.75}{
\begin{tabular}{|lll|l|l|}
\hline
   Eval   & Model      &           \textbf{dev} &           \textbf{test} \\
\hline
\textbf{Morph-} & \textit{morph gold} &  $\textit{80.03} \pm \textit{0.4}$ &  $\textit{79.10} \pm \textit{0.6}$ \\
\textbf{Level} &       \textbf{morph hybrid} &  $\textbf{78.51} \pm 0.5$ &  $\textbf{77.11} \pm 0.7$ \\
 &       \textbf{morph standard} &  $72.79 \pm 0.5$ &  $69.52 \pm 0.6$ \\
%      & \textbf{token-multi} & \textbf{gold} &  $77.48 \pm 0.5$ &  $77.04 \pm 0.3$ \\
      & \textbf{token-multi hybrid}  &  $75.70 \pm 0.5$ &  $74.64 \pm 0.3$ \\
\hline
\hline
\textbf{Token-} & \textit{morph gold} &  $\textit{80.30} \pm \textit{0.5}$ &  $\textit{79.28} \pm \textit{0.6}$ \\
\textbf{Level}      &  \textbf{morph hybrid} &  $\textbf{79.04} \pm 0.5$ &  ${77.64} \pm 0.7$ \\
 &       \textbf{morph standard} &   {74.52} $\pm$ {0.7} &  $73.53 \pm 0.8$ \\
      & \textbf{token-multi}        &  $77.59 \pm 0.4$ &  $\textbf{77.75} \pm 0.3$ \\
      & \textbf{token-single}        &  $78.15 \pm 0.3$ &  $77.15 \pm 0.6$ \\
\hline
\end{tabular}
 }
\caption{\label{tab:test-set-results} 
Test vs.\ Dev: Results with fastText for all Models. {\em morph-gold}  presents an ideal upper-bound. 
  }
  \vspace{-0.1in}
\end{table}

\subsection{Setting in the Greater Context}

\paragraph{Test Set Results}
Table \ref{tab:test-set-results}  confirms our best  results on the Test set. 
The  trends are kept,  though  results on Test are lower than on  Dev. The {\em morph gold} scenario still provides an upperbound of the performance, but it is not realistic. For the realistic scenarios, {\em morph hybrid} generally outperforms all other alternatives. The only divergence is that in token-level evaluation, \MULMACRO performs on a par with {\em morph hybrid} on the Test set.

\paragraph{Results on MD Tasks.} 
While the \FLIPMACRO pipeline achieves superior performance on NER,
it also improves the state-of-the-art on other  tasks in the pipeline. Table \ref{tab:seg-pos} shows the   Seg+POS results  of our \FLIPMACRO pipeline scenario, compared with the \YAPMACRO pipeline which replicates the pipeline of \citet{more-etal-2019-joint}. We use the  metrics defined by \citet{more-etal-2019-joint}. 
We show substantial improvements for the \FLIPMACRO pipeline
over the results of \citet{more-etal-2019-joint}, and    also outperforming the Test results of \citet{seker-tsarfaty-2020-pointer}.

\begin{table}[t]
\scalebox{0.8}{
\begin{tabular}{|ll|l|}

\hline
& &            \textbf{Seg+POS} \\%&            \textbf{Seg} &            \textbf{POS} \\

\hline
\textbf{dev} & \YAPMACRO \cite{more-etal-2019-joint} &  $92.36$ \\%&  $96.12$ &  $92.76$ \\
 & Ptr-Network \cite{seker-tsarfaty-2020-pointer} &  \textbf{93.90}\\ %&  &  \\
 & \FLIPMACRO (This work) &  $93.12$ \\%&  $\textbf{97.65}$ &  $\textbf{93.86}$ \\
\hline \hline
\textbf{test} & \YAPMACRO \cite{more-etal-2019-joint} &  $89.08$ \\ %&  $95.51$ &  $90.85$ \\
 & Ptr-Network \cite{seker-tsarfaty-2020-pointer} &  $90.49$ \\% &   &   \\

 & \FLIPMACRO (This work) &  $\textbf{90.89}$ \\% &  $\textbf{97.05}$ &  $\textbf{91.77}$ \\

\hline
\end{tabular}
}
\caption{\label{tab:seg-pos} 
Morphological Segmentation \& POS scores. %\citet{more-etal-2019-joint} and \citet{seker-tsarfaty-2020-pointer}.
  }
\end{table}

\begin{table}[t]
\centering
\scalebox{0.75}{
\begin{tabular}{|l|l|l|l|}
\hline
 & \textbf{Precision} & \textbf{Recall} & \textbf{F1}\\   
\hline
\newcite{naama}  & $84.54$ & $74.31$ &  $79.10$\\
MEMM+HMM+REGEX & & &\\
\hline
This work & $\text{86.84}$ &  $\text{82.6}$ & $\text{84.71}$\\
\TOKMACRO+FT+CharLSTM & $ \pm 0.5$ & $ \pm 0.9$ & $ \pm 0.5$\\
\hline
This work & $\textbf{86.93}$ &  $\textbf{83.59}$ & $\textbf{85.22}$\\
\emph{morph}-\FLIPMACRO+FT+CharLSTM & $ \pm 0.6$ & $ \pm 0.8$ & $ \pm 0.5$\\

\hline
\end{tabular}
 }
\caption{\label{tab:naama-results} 
NER Comparison with \citet{naama}.
  }
  \vspace{-0.05in}
\end{table}
\paragraph{Comparison with Prior Art.} 
Table \ref{tab:naama-results} presents our results on the Hebrew NER corpus of \citet{naama} compared to their model, which uses a hand-crafted feature-engineered MEMM  with 
regular-expression rule-based enhancements and an entity lexicon. 
Like \citet{naama} we performed three 75\%-25\% random train/test splits,
and used the same seven NE categories {\sc (Per,Loc,Org,Time,Date,Percent,Money)}.
We trained a \TOKMACRO model on the original space-delimited tokens and a \MORMACRO model on automatically segmented morphemes we obtained using our best segmentation model (\FLIPMACRO MD on our trained \MULMACRO model, as in Table \ref{tab:seg-pos}). Since their annotation includes only token-level boundaries, all of the results we report conform with token-level evaluation.     

Table \ref{tab:naama-results} presents the results of these experiments. Both  models significantly outperform the previous state-of-the-art by \citet{naama}, setting a new performance bar on this earlier benchmark. Moreover, we again observe an empirical advantage when explicitly modeling morphemes, {\em even} with the automatic noisy segmentation that is used for the morpheme-based training.

\section{Discussion: Joint Modeling Alternatives and Future Work} 
The present study provides the motivation and the necessary foundations  for comparing morpheme-based and token-based modeling for NER. 
While our findings  clearly demonstrate the advantages of morpheme-based modeling for NER in a morphologically rich language, it is clear  that our proposed {\em Hybrid} architecture is not the only modeling alternative for linking NER and morphology. 

For example, a previous study by \citet{DBLP:gungor-joint-md-ner} addresses {\em joint} neural modeling of morphological segmentation and NER labeling,   proposing a multi-task learning (MTL) approach for joint MD and NER in Turkish. They employ separate Bi-LSTM networks for the MD and NER tasks, with a shared loss to allow for joint learning. Their results indicate improved NER performance, with no improvement in the MD results. 
Contrary to our proposal, they view MD and NER as distinct tasks,  assuming a single NER label per token, and not providing disambiguated morpheme-level boundaries for the NER task. More generally, they test only {\em token-based} NER labeling   and do not attend to the question of input/output  granularity in their models.

A  different  approach  for  joint  NER  and  morphology is jointly predicting the segmentation and labels for each token in the input stream. This is the approach taken, for instance, by the lattice-based Pointer-Network of \citet{seker-tsarfaty-2020-pointer}. As shown in Table 5, their results for morphological segmentation  and  POS  tagging  are  on  a par  with our  reported  results  and,  at  least  in  principle, it should be possible to extend the \citet{seker-tsarfaty-2020-pointer} approach to yield also NER predictions. 

However, our preliminary experiments with a lattice-based Pointer-network for token segmentation and NER labeling shows that this is not a straightforward task. Contrary to POS tags, which are constrained by the MA, every NER label can potentially go with any  segment, and this leads to a combinatorial explosion of the search space represented by the lattice. As a result, the NER predictions are brittle to learn, and the complexity of the resulting model is computationally prohibitive.

A different approach to joint sequence segmentation and labeling can be  applying the neural model directly on the character-sequence of the input stream. 
Such an approach is for instance the char-based {\em labeling as segmentation} setup  proposed by \citet{shao-etal-2017-character}. Shao et al.\  use a character-based Bi-RNN-CRF to output a single label-per-char which indicates  both  word boundary (using BIES sequence labels) and the POS tags.
This method is also used in their universal segmentation paper, \cite{shao-etal-2018-universal}.
However, as seen in the results of \citet{shao-etal-2018-universal}, char-based labeling for segmenting Semitic languages lags far behind all other languages, precisely because morphological boundaries are not explicit in the character sequences. 

Additional proposals are those of \citet{kong2015segmental,kemos-etal-2019-neural}.
First, \citet{kong2015segmental} proposed to solve
 e.g.\ Chinese segmentation and POS tagging using
dynamic programming with neural encoding, by using a Bi-LSTM to encode the character input, and then feed it to a semi-markov CRF to obtain probabilities for the different segmentation options. 
\citet{kemos-etal-2019-neural} propose an approach similar to \citet{kong2015segmental} for joint segmentation and tagging but add convolution layers on top of the Bi-LSTM encodings to obtain segment features hierarchically and then feed them to the semi-markov CRF.

Preliminary experiments we conducted confirm that  char-based  joint segmentation and NER labeling for Hebrew, either using char-based labeling or a seq2seq architecture, still lags  behind our reported results. We conjecture that this is due to the complex morpho-phonological and orthographic processed in Semitic languages. Going into char-based modeling nuances and offering a sound joint solution for a language like Hebrew is an important matter that merits its own investigation. Such work is feasible now given the new corpus, however, it is out of the scope of the current study.

All in all, the design of  sophisticated joint modeling strategies for morpheme-based NER poses fascinating questions --- for which our work provides a solid  foundation  (data, protocols, metrics, strong baselines).
More work is needed for investigating joint modeling of  NER and morphology, in the directions portrayed in this Section, yet it is beyond the scope of this  paper, and we leave this investigation for future work. 

Finally, while the joint approach %alluded to above 
is appealing, we argue that the elegance of our  {\em Hybrid} solution is precisely in providing a clear and well-defined interface between MD and NER through which the two tasks can interact, while still keeping the distinct models simple, robust, and efficiently trainable. It also has the advantage of allowing us to seamlessly integrate sequence labelling with any  lattice-based MA,
in a plug-and-play language-agnostic  fashion, towards obtaining further advantages on both of these tasks.

\section{Conclusion}
\label{sec:conclusions}
This work addresses the modeling challenges of Neural NER in MRLs. 
We deliver a  parallel {\em token-vs-morpheme} NER corpus for Modern Hebrew, that allows one to assess   NER modeling strategies in morphologically rich-and-ambiguous environments. Our experiments show that while 
NER   benefits from morphological decomposition,
 downstream results are  sensitive to segmentation errors.
We thus  propose a   \FLIPMACRO architecture in which NER {\em precedes} and  {\em prunes} the morphological decomposition. This approach greatly outperforms a \YAPMACRO~ {pipeline} in realistic (non-gold)  scenarios.
Our analysis further shows  that morpheme-based models better recognize OOVs that result from morphological composition. 
All in all we deliver new state-of-the-art results for  Hebrew NER and MD,  along with a novel benchmark, to encourage further investigation  into the interaction between  NER and morphology.

\section*{Acknowledgments}
We are grateful to the BIU-NLP lab members as well as 6 anonymous reviewers for their insightful remarks. We further thank Daphna Amit and Zef Segal for their meticulous annotation and profound discussions. This research is funded by an ISF Individual Grant (1739/26) and an ERC Starting Grant (677352), for which we are grateful.

\bibliography{tacl_ner}
\bibliographystyle{acl_natbib}

\appendix

\section{Alignment Heuristics}
\label{sec:appendix-extending-morph}

{\bf Aligning Multi-labels to Single Labels.}
In order to evaluate morpheme-based labels (morph or token-multi) in token-based settings, we introduce a deterministic procedure to extend the morphological labels to token boundaries.
Specifically, we use regular expressions to map the multiple sequence labels  to a single label by choosing the first non-O entity category (BIES) as the single category. In case the sequence of labels is not valid (e.g., B comes after E, or there is an O between two I labels), we use a relaxed mapping that does not take the order of the labels into consideration: if there is an S or {\em both} B and E in the sequence, return an S. Otherwise, if there is an E, return an E; if there is a B, return a B; if there is an I return an I (Figure~\ref{appendix:align-to-labels}).
\\\\

\noindent
{\bf Aligning Multi-labels to Morphemes.}
In order to obtain morpheme boundary labels from \emph{token-multi}, we introduce a deterministic procedure to align the token's predicted multi-label with the list of  morphemes predicted for it by the MD. 
Specifically, we  align the multi-labels to  morphemes in the order that they are both provided. In  case of a mismatch between  the number of labels and morphemes predicted for the token, we match  label-morpheme pairs from the final one  backwards.  If the number of morphemes exceeds the number of labels, we pad unpaired morphemes with O labels. If the 
%In case the 
number of labels exceeds the   morphemes, we drop  unmatched labels (Figure~\ref{appendix:align-to-morphs}).

\begin{figure}[t]
\textbf{def extend\_morph\_to\_tok(token\_labels):}\\
\emph{\# gets list of morpheme NER labels for a token}\\
\emph{\# returns single NER label}\\
\indent biose\_labels = ''\\
\indent categories = list()\\
\indent for label in token\_labels:\\
\indent \indent if label=='O': \\
\indent \indent \indent biose\_labels += 'O' \\
\indent \indent \indent categories += None \\
\indent \indent else: \\
\indent \indent \indent bio, cat = label.split('-') \\
\indent \indent \indent biose\_labels += bio\\
\indent \indent \indent categories += cat\\
\indent single\_biose = get\_single\_biose(biose\_labels)\\
\indent single\_cat = get\_single\_token\_categ(categories)\\
\indent if single\_biose == 'O':\\
\indent \indent return 'O'\\
\indent else:
\indent return single\_biose + '-' +single\_cat \\\\
\textbf{def get\_single\_token\_categ(categ\_labels):}\\ 
\emph{\# gets list of NER categories for a token}\\
\emph{\# returns the first category which is not None }\\
\indent [...]\\\\
\textbf{def get\_single\_biose(biose\_labels):}\\ 
\emph{\# gets string of 'BIOSE' for a token}\\
\emph{\# returns single character from 'BIOSE'}\\
\indent \emph{\# regex matches valid labels for: O|B|S|I|E|S}
\indent valid\_seq = {\footnotesize 'O+|O*BI*|O*BI*EO*|I+|I*EO*|O*SO*'}\\
\indent if valid\_seq.match(biose\_labels):\\
\indent \indent \emph{\# split valid\_seq by '|' and return} \\
\indent \indent \emph{\# label based on the corresponding re match } \\
\indent \indent [...] \\ 
\indent else: \emph{\# treat as set}\\
\indent \indent if 'S' in biose\_labels:\\
\indent \indent\indent return 'S'\\
\indent \indent    elif 'B' and 'E' in biose\_labels:\\
\indent \indent\indent        return 'S'\\
\indent \indent    elif 'E' in biose\_labels:\\
\indent \indent\indent        return 'E'\\
\indent \indent    elif 'B' in biose\_labels:\\
\indent \indent\indent        return 'B'\\
\indent \indent    elif 'I' in biose\_labels:\\
\indent \indent\indent        return 'I'\\
\indent \indent    return 'O'\\
\caption{Multi-Label to Single Label Alignment }\label{appendix:align-to-labels}
\end{figure}

\begin{figure}[t]
\textbf{def align\_multi\_to\_morph(labels, forms):}\\
\emph{\# gets list of labels and list of morph forms}\\
\emph{\# returns aligned list of tuples (morph, label)}\\
\indent if len(labels)==len(forms):\\ 
\indent \emph{\# if same length, return zipped by order}\\
\indent \indent return zip(forms, labels)\\
\indent elif len(labels)>len(forms):\\
\indent \emph{\# if extra labels, trim labels from the beginning}\\
\indent \emph{\# return zipped by order}\\
\indent \indent [ ... ]\\
\indent elif len(labels)<len(forms):\\
\indent \emph{\# if extra forms, pad beginning with 'O'}\\
\indent \emph{\# return zipped by order}\\
\indent \indent [ ... ]

    \caption{Multi-Label to Morpheme Alignment}
    \label{appendix:align-to-morphs}
\end{figure}
\end{document}